\documentclass[conference]{IEEEtran}
\IEEEoverridecommandlockouts
\usepackage{cite}
\usepackage{amsmath,amssymb,amsfonts}
\usepackage{algorithmic}
\usepackage{graphicx}
\usepackage{textcomp}
\usepackage{subcaption}
\usepackage{caption}
\usepackage[utf8]{inputenc}
\usepackage{multirow}
\usepackage{array}
\usepackage{float} 
\usepackage{booktabs}
\usepackage{tabularx}
\usepackage[table]{xcolor}

\def\BibTeX{{\rm B\kern-.05em{\sc i\kern-.025em b}\kern-.08em
    T\kern-.1667em\lower.7ex\hbox{E}\kern-.125emX}}
\definecolor{lightyellow}{RGB}{255,255,200}
\definecolor{lightred}{RGB}{255,200,200}

\begin{document}

\title{\\
}

		{

\title{AquaGS: Fast Underwater Scene Reconstruction with SfM-Free Gaussian Splatting\\
}

\author{\IEEEauthorblockN{ Junhao Shi}
	\IEEEauthorblockA{\textit{School of Electronic Information and Electrical Engineering} \\
		\textit{Shanghai  Jiao  Tong University}\\
		Shanghai, China \\
		stone.xx@sjtu.edu.cn}
	\and
	\IEEEauthorblockN{ Jisheng Xu}
	\IEEEauthorblockA{\textit{School of Electronic Information and Electrical Engineering} \\
		\textit{Shanghai  Jiao  Tong University}\\
		Shanghai, China \\
		Jimmy\_xu@sjt.edu.cn}
	\and
	\IEEEauthorblockN{ Jianping He}
	\IEEEauthorblockA{\textit{School of Electronic Information and Electrical Engineering} \\
		\textit{Shanghai  Jiao  Tong University}\\
		Shanghai, China \\
		jphe\@sjtu.edu.cn}
}

\author{\IEEEauthorblockN{\textsuperscript{1}Junhao Shi, \textsuperscript{2}Jisheng Xu, \textsuperscript{3}Jianping He, \textsuperscript{4}Zhiliang Lin}
	\IEEEauthorblockA{\textit{\textsuperscript{1,2,3}School of Electronic Information and Electrical Engineering, \textsuperscript{4}State Key Laboratory of Ocean Engineering} \\
		\textit{Shanghai  Jiao  Tong University}\\
		Shanghai, China \\
		\textsuperscript{1}stone.xx@sjtu.edu.cn, \textsuperscript{2}Jimmy\_xu@sjtu.edu.cn, \textsuperscript{3}jphe@sjtu.edu.cn,
		\textsuperscript{4}linzhiliang@sjtu.edu.cn}
}

\maketitle

\begin{abstract}
Underwater scene reconstruction is a critical technology for underwater operations, enabling the generation of 3D models from images captured by underwater platforms. However, the quality of underwater images is often degraded due to medium interference, which limits the effectiveness of Structure-from-Motion (SfM) pose estimation, leading to subsequent reconstruction failures. Additionally, SfM methods typically operate at slower speeds, further hindering their applicability in real-time scenarios. In this paper, we introduce AquaGS, an SfM-free underwater scene reconstruction model based on the SeaThru algorithm, which facilitates rapid and accurate separation of scene details and medium features. Our approach initializes Gaussians by integrating state-of-the-art multi-view stereo (MVS) technology, employs implicit Neural Radiance Fields (NeRF) for rendering translucent media and utilizes the latest explicit 3D Gaussian Splatting (3DGS) technique to render object surfaces, which effectively addresses the limitations of traditional methods and accurately simulates underwater optical phenomena. Experimental results on the data set and the robot platform show that our model can complete high-precision reconstruction in 30 seconds with only 3 image inputs, significantly enhancing the practical application of the algorithm in robotic platforms.
\end{abstract}
\section{Introduction}

Underwater scene reconstruction refers to create a three-dimensional representation of underwater environments from a series of two-dimensional images. This process is essential for various applications, including marine archaeology, environmental monitoring, and underwater exploration \cite{huOverviewUnderwater3D2023}.

However, significant challenges arise due to the degradation of underwater images caused by absorption and scattering. As light travels through water, it is scattered or absorbed by particles, weakening the signals reaching the image sensors. Additionally, light at different wavelengths decays at varying rates, leading to the characteristic blue-green tint and haziness of underwater images \cite{schechnerClearUnderwaterVision2004a}. This degradation undermines the accuracy of underwater scene reconstruction.

Existing scene reconstruction models, although effective in terrestrial environments, struggle to adapt to underwater conditions. Their reliance on precise initial camera parameter estimations via Structure-from-Motion (SfM)~\cite{wangDUSt3RGeometric3D2023} proves problematic. The SfM process is both computationally expensive and unstable. Especially in underwater settings, the degradation of underwater images often leads to a lack of sufficiently clear feature points in the input data, which causes cumulative errors during feature extraction and matching. As a result, accurate camera parameter estimation becomes challenging, and in some cases, reconstruction fails entirely~\cite{fanInstantSplatSparseviewSfMfree2024}. Moreover, even with accurate camera pose estimation, current models fail to incorporate the medium's effects based on the principles of underwater image formation, further hindering accurate restoration of underwater objects.

To address the limitations of SfM-based models, InstantSplat was proposed \cite{fanInstantSplatSparseviewSfMfree2024}, utilizing the pre-trained DUSt3R model \cite{wangDUSt3RGeometric3D2023} for efficient multi-view stereo point cloud generation. However, it is unsuitable for underwater scenes. To efficiently account for the effects of the underwater medium, SeaThru-NeRF\cite{levySeaThruNeRFNeuralRadiance2023} was introduced to separately model objects and the medium, aiming to restore the true appearance of underwater environments. Nevertheless, SeaThru-NeRF is hampered by slow rendering and training performance.

In this paper, we propose a novel underwater scene reconstruction model named \textbf{AquaGS}. By utilizing a pre-trained neural network model, we can rapidly and stably generate an initial point cloud, thus initializing Gaussians. The 3D Gaussian Splatting (3DGS) component of AquaGS effectively captures surface information of underwater objects, while the neural radiance field part models the underwater medium. The integrated framework allows for precise modeling of both object geometry and the complex optical properties of underwater scenes, enabling the generation of novel underwater views in seconds and effectively mitigating the impact of the medium between the camera and the scene. Our main contributions are as follows:

\begin{itemize}
	\item \textbf{Stable Initialization:} We obtain a more stable gaussian initialization with a pre-trained neural network, which significantly mitigates the impact of image degradation.
	\item \textbf{Removal of Medium:} We assign distinct parameters to underwater objects and the medium, utilizing Gaussian splatting and neural radiance fields to achieve precise reconstruction of underwater scenes.
	\item \textbf{Real-time Training and Rendering:} We validate the efficiency and effectiveness of AquaGS through extensive dataset testing and deployment on underwater robotic platforms. Our model facilitates high-quality 3D underwater reconstruction in just seconds.
\end{itemize}

\section{Related Work}

\subsection{\textbf{3D Reconstruction Preprocessing}}

Image-based unconstrained 3D reconstruction is a key objective in computer vision. Many algorithms, including SeaThru-NeRF~\cite{levySeaThruNeRFNeuralRadiance2023}, 3DGS~\cite{kerbl3DGaussianSplatting2023}, WaterSplatting\cite{liWaterSplattingFastUnderwater2024},  depend on accurate camera pose estimation, typically achieved through Structure-from-Motion (SfM) software like COLMAP~\cite{fanInstantSplatSparseviewSfMfree2024}. However, the reliance on SfM limits practical applications. For instance, COLMAP requires high-quality images with sufficient overlap and rich textures for successful feature matching and reconstruction~\cite{lindenbergerPixelPerfectStructureMotionFeaturemetric}. This becomes challenging in underwater environments where image degradation often impairs these methods. Additionally, COLMAP’s time-consuming computation, which can take hours per scene, makes it unsuitable for real-time underwater robotic platforms.

To overcome these limitations, recent methods directly predict 3D structures from a single RGB image. Advances in deep learning-based dense stereo frameworks have reduced scale-aware depth map inference to milliseconds \cite{guCascadeCostVolume2020,yaoMVSNetDepthInference2018}. These methods leverage neural networks to learn 3D priors from large datasets, often adapting Monocular Depth Estimation networks to generate depth maps. By incorporating camera intrinsics, they can produce pixel-aligned 3D point clouds. For instance, SynSin~\cite{wilesSynSinEndEndView2020} generates novel viewpoints from a single image by rendering feature-enhanced depth maps with known camera parameters. However, it is inherently limited by the quality of the depth estimate, which may not be appropriate for a monocular setting. In contrast, DUSt3R~\cite{wangDUSt3RGeometric3D2023} processes multiple viewpoints to generate depth maps or point maps without requiring ground-truth camera intrinsics, making it applicable to diverse scenes and configurations, thus overcoming many limitations of traditional SfM methods.

\subsection{\textbf{Physics-Based Models for Underwater Image Formation}}

The attenuation of underwater light is influenced not only by the inherent properties of water but also by factors such as camera sensors, scene structure, and material characteristics \cite{mcglameryComputerModelUnderwater1980,jaffeComputerModelingDesign1990}. Although ocean optical instruments like transmissometers and spectrometers can measure attenuation coefficients, differences in spectral sensitivity and acceptance angles limit their direct applicability to imaging \cite{bongiornoDynamicSpectralbasedUnderwater2013a}.

A research direction in underwater image formation models is based on physical principles, typically starting from the radiative transfer equation \cite{chandrasekharRadiativeTransfer1960}. However, fully solving this equation requires extensive Monte Carlo simulations, which are unsuitable for real-time rendering. As a result, many methods simplify the problem by incorporating priors, such as the Dark Channel Prior \cite{carlevaris-biancoInitialResultsUnderwater2010} and Haze Line Prior \cite{bermanNonlocalImageDehazing2016}, to separate backscattering and transmitted components. Nonetheless, these approaches often rely on stereo cameras and prior knowledge of the water type.

To address the complexity of underwater medium parameters, the SeaThru model was introduced~\cite{akkaynakWhatSpaceAttenuation2017,akkaynakRevisedUnderwaterImage2018}. This model decomposes underwater image formation into three components: direct signals, backscatter, and forward-scattering. Direct signals are reflections from the scene, while backscatter consists of stray light entering the camera, often introducing noise. Forward-scattering, caused by light deviating from its path, minimally impacts image degradation~\cite{mobleyLightWaterRadiative,schechnerClearUnderwaterVision2004a}. The mathematical model of underwater image formation is expressed as:

\begin{equation}
I_\text{R,G,B} = D_\text{R,G,B} \cdot f(\beta^D) + B_\text{R,G,B} \cdot f(\beta^B),
\end{equation}
where \( I_\text{R,G,B} \) is the captured image, \( D_\text{R,G,B} \) is the direct signal, and \( B_\text{R,G,B} \) is the backscatter signal, both modulated by their respective attenuation factors \( f(\beta^D) \) and \( f(\beta^B) \)~\cite{akkaynakWhatSpaceAttenuation2017,akkaynakRevisedUnderwaterImage2018}.

\begin{figure*}[h]  
	\centering
	\includegraphics[width=\linewidth]{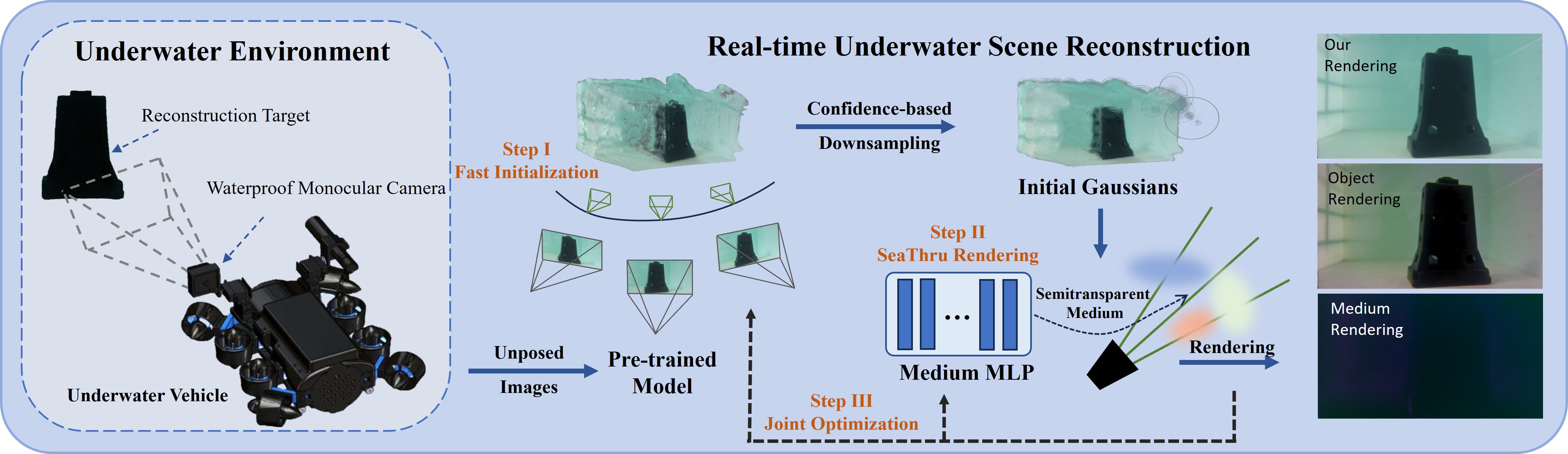}
	\caption{\textbf{Overview of AquaGS. }In the underwater environment, our vehicle captures real-time images, sequentially performing Fast Initialization, SeaThru Rendering and Joint Optimization, producing an accurate restoration of underwater objects.
	}
	\vspace{-10pt}
\end{figure*}

\subsection{\textbf{NeRF and 3D Gaussian Splatting}}
Neural radiance field (NeRF)~\cite{mildenhallNeRFRepresentingScenes2021} technology has seen widespread use in scene reconstruction. It represents 3D scenes using multi-layer perceptrons (MLPs), taking the direction and position of 3D points as inputs and synthesizing images by performing volumetric rendering integration over a series of samples along a ray. In terms of eliminating the need for SfM, NeRFmm~\cite{wangNeRFNeuralRadiance2022} jointly optimizes camera parameters and NeRF by treating them as learnable variables through photometric reconstruction. NoPe-NeRF~\cite{bianNoPeNeRFOptimisingNeural2023} utilizes an undistorted monocular depth prior to constrain relative poses, allowing joint optimization of camera poses and NeRF in large-motion sequences;
In terms of underwater 3D reconstruction, WaterNeRF~\cite{sethuramanWaterNeRFNeuralRadiance2023} employs underwater images and their histogram-equalized versions to estimate absorption and backscattering coefficients. SeaThru-NeRF~\cite{levySeaThruNeRFNeuralRadiance2023} builds on the SeaThru model by using two networks to separate true underwater images from backscatter, removing scattering without prior color distribution knowledge.

However, NeRF's reliance on intensive neural network inference at multiple points along each ray makes real-time performance difficult to achieve. In contrast, 3DGS\cite{kerbl3DGaussianSplatting2023} simplifies 3D scene representation using Gaussian functions, enabling rendering speeds of over 100 fps while maintaining differentiability, significantly reducing computational complexity and rendering time. Therefore, we enhanced 3DGS by integrating learning-based MVS technology and an underwater imaging model, creating the AquaGS algorithm, which improves underwater perception in 3D reconstruction.

\section{Algorithm Design}
As shown in Fig. 1, the workflow of the proposed AquaGS model consists of three steps. \textbf{Step I: Fast Initialization}, we utilize the pre-trained DUSt3R model to generate dense, pixel-aligned point sets. Subsequently, farthest confidence-based downsampling is applied to retain high-confidence Gaussians, simplifying training and providing a refined initialization. \textbf{Step II: SeaThru Rendering}, explicit Gaussian Splatting captures surface geometry, while implicit Neural Radiance Fields learn the medium's transmittance and render the segments between Gaussians. \textbf{Step III: Joint Optimization}, photometric loss is used to jointly optimize camera parameters and scene attributes, enabling fast, high-fidelity novel view synthesis.

\subsection{\textbf{Incorporating Learning-Based MVS}}
To overcome the limitations of SfM-based methods, a direct approach is to map RGB images to dense point clouds. We introduce an innovative paradigm for unconstrained stereo 3D reconstruction using DUSt3R, enabling 3D model generation from arbitrary image sets without the need for prior camera calibration or viewpoint poses. Specifically, our model regresses unprojected, normalized point maps from two input views, with the regression loss defined as:
\vspace{-3pt}
\begin{equation}
	\mathcal{L}_{\text{reg}} = \left\| \frac{1}{s_i} M_v - \frac{1}{\hat{s}_i} \hat{M}_v \right\|, \nonumber
\end{equation}
here, \(M_v\) is the predicted point map, and \(\hat{M}_v\) is the ground truth with view \(v \in \{1, 2\} \). The \(s_i\) and \(\hat{s}_i\) are corresponding scaling factor to handle scale ambiguity by normalizing both predicted and ground-truth point maps. Additionally, DUSt3R defines a pixel-wise confidence and optimizes it using a confidence loss:
\begin{equation}
	\mathcal{L}_{\text{conf}} = \sum_{v \in \{1,2\}} \sum_{i \in D_v} \left[ \gamma_i^v \cdot \mathcal{L}_{\text{reg}}(v, i) - \alpha \cdot \log \gamma_i^v \right],\nonumber
\end{equation}
here, the hyperparameter \(\alpha\) controls regularization, encouraging the network to focus on more challenging areas by extrapolating where confidence is lower.


The original DUSt3R takes a single image pair as input. When more than two views are captured, post-processing is needed to align scales. By constructing a fully connected graph, where nodes represent input views and edges represent image pairs with shared content, we optimize the point cloud map, transformation matrix, and scale factor to align the normalized point clouds into a common coordinate system.

To address the redundancy in DUSt3R’s dense point clouds, a confidence-based voxel grid downsampling strategy is employed. The point cloud is divided into uniform voxel grids, with the most representative point in each voxel selected based on its confidence score. This ensures uniform scene coverage while reducing redundant data, improving rendering efficiency.

After downsampling, the dense point cloud is converted into a Gaussian point set. Each point is represented as a Gaussian characterized by its position \( \mathbf{x} \), spherical harmonic coefficients \( sh_i \), opacity \( \alpha_i \), rotation \( q_i \), and scaling \( s_i \). 
\subsection{\textbf{SeaThru Splatting Rendering}}

After initializing Gaussian with efficient learning-based MVS technology, we use Gaussian and neural radiation fields to render underwater scenes. 
During the rendering process, the color of each pixel \( p \) in the image is derived by combining \( N \) sequentially ordered Gaussians \( \{G_i | i = 1, \ldots, N\} \) that overlap at \( p \), as follows:
\begin{equation}
	C(p) = \sum_{i=1}^{N} \left( c_i \alpha_i \prod_{j=1}^{i-1} (1 - \alpha_j) \right)\nonumber,
\end{equation}
where, \( \alpha_i \) is the product of the 2D Gaussian value from \( G_i \) at pixel \( p \) and the learned opacity for \( G_i \), and \( c_i \) is the adjustable color parameter of \( G_i \). Gaussians overlapping at \( p \) are ordered by depth from the viewer's perspective.

Inspired by 3DGS and SeaThru model (Eq.1), our model uses separate color and density parameters for the objects and medium:
\begin{equation}
	C(p) = \sum_{i=1}^{N} \left[ \left({c_i^\text{obj}} {\alpha_i^\text{obj}} + {c_i^\text{med}} {\alpha_i^\text{med}}\right) \prod_{j=1}^{i-1} (1 - \alpha_j) \right],\nonumber
\end{equation}

In addition, we assume that the object in the scene is opaque, while the medium is translucent, characterized by a lower non-zero density. So that $\alpha_i^{\text{med}} \gg \alpha_i^{\text{obj}}$ before the object and $\alpha_i^{\text{med}} \ll \alpha_i^{\text{obj}}$. Based on this, we use Gaussians and NeRF to render both the objects and the medium.
\subsection{\textbf{Loss Function Setting}}
In the SeaThru-NeRF~\cite{levySeaThruNeRFNeuralRadiance2023}, the loss function consists of three parts, making it suitable for separating the medium from the object. Inspired by this, we compute the loss function through several components. We utilize the reconstruction loss \( \mathcal{L}_{\text{recon}} \) to measure the difference between the generated image \( \hat{C} \) and the ground truth image \( C^* \). This loss consists of two parts: 
\begin{equation}
	\begin{array}{c}
		\mathcal{L}_{\text{recon}}^{\text{L1}} = |\hat{C} - C^*|,\quad
		\mathcal{L}_{\text{recon}}^{\text{D-SSIM}} = \text{D-SSIM}(\hat{C}, C^*), \\[0.5em]
		\mathcal{L}_{\text{recon}} = \mathcal{L}_{\text{recon}}^{\text{L1}} + \mathcal{L}_{\text{recon}}^{\text{D-SSIM}},
	\end{array}
	\nonumber
\end{equation}
the L1 Loss captures the absolute pixel-wise difference and the D-SSIM Loss evaluates the structural similarity between the generated image and the ground truth. 

Additionally, we introduce the transparency loss \( \mathcal{L}_{\text{acc}} \) to ensure that the transparency values of the sampled points are constrained to either 0 or 1. This involves modeling a prior distribution for transparency as a mixture of two Laplacian distributions given by:
\begin{equation}
	\begin{array}{c}
		P(T_{\text{obj}}) \propto e^{-|T_{\text{obj}}|/0.1} + e^{-|1 - T_{\text{obj}}|/0.1}, \\[0.4em]
		\mathcal{L}_{\text{acc}} = -\log P(T_{\text{obj}}).
	\end{array}
	\nonumber
\end{equation}

Finally, the overall loss function \( \mathcal{L} \) is formulated by combining these three components, resulting in:
\begin{equation}
	\mathcal{L} = \mathcal{L}_{\text{recon}} + \lambda \mathcal{L}_{\text{acc}},\nonumber
\end{equation}
where \( \lambda \) is a hyperparameter that adjusts the weight of the transparency loss in the total loss. By optimizing this loss function, we can effectively improve the model's performance and ensure that the generated images better match the features of the real images.

\subsection{\textbf{Algorithm Implementation}}
Our implementation is based on the original 3DGS  code~\cite{kerbl3DGaussianSplatting2023}. For the point cloud initialization part, we adjust the input image resolution to 512 and then enter DUSt3R. For the splatting part, we disable the adaptive density control of the Gaussian ellipsoid due to the high quality of the initial point cloud. For the Medium MLP, we use 2 linear layers with 128 features and a Sigmoid activation, followed by Sigmoid activation for predicting \(c^\text{med}\).

\section{Experiments and Results}

\subsection{\textbf{Experimental Setup}}
\noindent
\textbf{SeaThruNeRF Dataset.} Our test dataset is sourced from the SeaThruNeRF Dataset, which includes four ocean scenarios: Curaçao, IUI3 Red Sea, Japanese Gardens Red Sea, and Panama~\cite{levySeaThruNeRFNeuralRadiance2023}, containing 29, 21, 20, and 18 images respectively. The dataset captures diverse water and imaging conditions. RAW images were taken using a Nikon D850 SLR camera housed in a Nauticam underwater housing with domed ports, avoiding refraction issues with the pinhole model. Images were downsampled to approximately 900 × 1400 pixels. White balance was applied to the linear images, with 0.5\% cropping per channel to eliminate extreme noise. Finally, COLMAP was used to extract camera poses.
\\
\begin{figure}[htbp]
	\centering
	\begin{minipage}{\linewidth}
		\centering
		\includegraphics[height=0.5\linewidth]{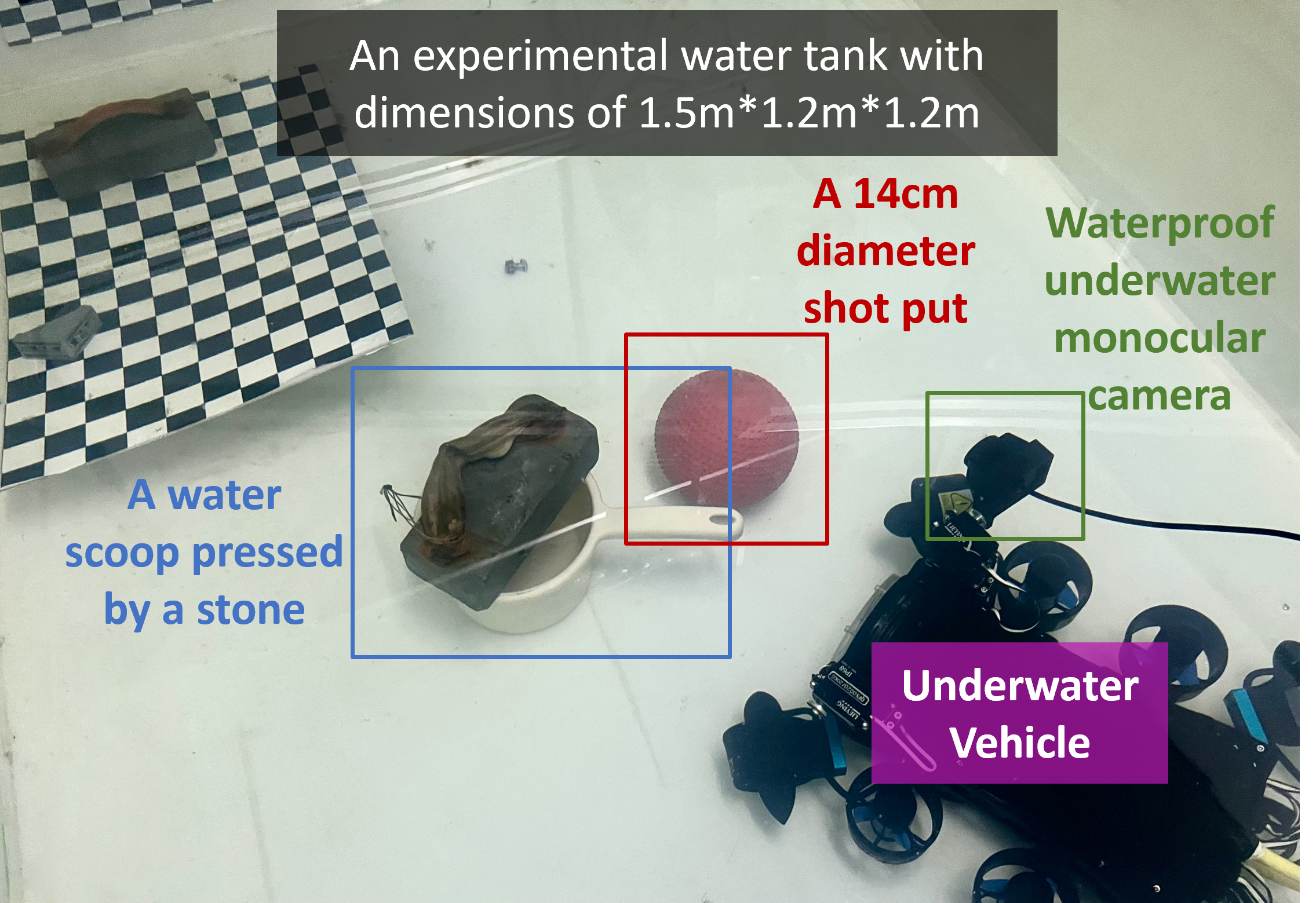}		
	\end{minipage}
	\caption{\textbf{Testing Bed Settings. }We integrate our model into robot. The reconstruction results are achieved on the computer.}
\end{figure}

\vspace{-5pt}\noindent
\textbf{Experiment with Underwater Robot. }To test the real-time performance and usefulness of our algorithm, we utilize a Remotely Operated Vehicle (ROV) to run tests on our AquaGS. The experiment took place in a water tank filled to a depth of approximately 50cm. In the first experiment, as shown in Fig. 1, we put a black barricade about 80cm high at the bottom; In the second experiment, we put a 14 diameter shot put and a water scoop pressed by a stone at the bottom of the tank (Fig. 2). Equipped with a waterproof monocular camera featuring a 2.7mm focal length and a 120° longitudinal field of view, the robot captured images (2592 × 1944 pixels) at a rate of 30 frames per second. The images are transmitted via cables to a ground-based computer, enabling fast 3D reconstruction.
\\

\vspace{-5pt}\noindent
\textbf{Baseline Methods. }We compare our method AquaGS with recent state-of-the-art 3D reconstruction methods, i.e. 3D Gaussian Splatting\cite{kerbl3DGaussianSplatting2023}, SeaThru-NeRF~\cite{levySeaThruNeRFNeuralRadiance2023} and InstantSplat~\cite{fanInstantSplatSparseviewSfMfree2024}. For our AquaGS and InstantSplat, we input the data directly into them; For 3DGS and SeaThru-NeRF, we utilize COLMAP for pre-computing the camera parameters. 
The optimization iterations for \textbf{Ours-S} are set to 500, while for \textbf{Ours-M} they are set to 1000, striking a balance between quality and training efficiency. In order to ensure the fairness of the test results, our training and rendering are carried out on the same Nvidia GeForce RTX 4090D GPU, and all experimental results were the average of multiple experiments.

\subsection{\textbf{Experimental Results}}
\noindent
\textbf{Dataset Results. }Fig. 3 shows the qualitative results of underwater scene rendering between AquaGS and various baseline methods on the standard benchmark dataset SeaThru-NeRF. Our AquaGS and SeaThru-NeRF achieve excellent results, while 3DGS and InstantSplat cannot effectively deal with the effects of underwater media on objects. 3DGS pruned Gaussians with low opacity, leaving dense and cloudy cloud-like primitives to fit the medium. InstantSplat, because it inputs sparse images, is prone to attempting aliasing, producing a lot of ghosting. 

As shown in Figure 4, our method performs well in reconstructing fine details (red squares). In the first three images, our model accurately predicts the scene's depth map (bottom left), while in the last two images, it effectively separates the medium (bottom left) from the objects, demonstrating the effectiveness of our approach.

TABLE I compares the PSNR\textsuperscript{$\uparrow$}, SSIM\textsuperscript{$\uparrow$}, LPIPS\textsuperscript{$\downarrow$}, and Avg. Training Time\textsuperscript{$\downarrow$} of AquaGS against baseline methods across four scenes. Our method achieved comparable results while demonstrating significant advantages in speed.
\\

\begin{figure*}[!hbtp] 
	\centering
	\includegraphics[width=\linewidth]{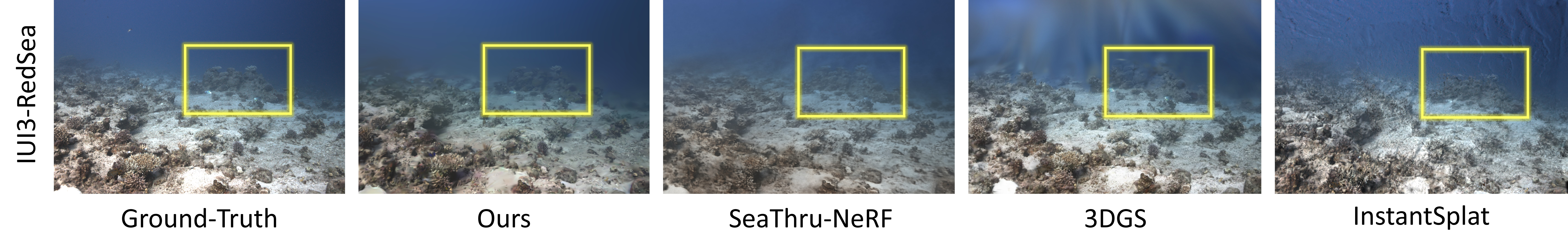}
	\caption{\textbf{Visual Comparisons of Underwater Scene Rendering between AquaGS and Various Baseline Methods. }AquaGS achieves excellent results, while 3DGS and InstantSplat cannot effectively deal with the effects of underwater media on objects.
	}
	\vspace{-5pt}
\end{figure*}
\begin{figure*}[!htbp]  
	\centering
	\includegraphics[width=1.01\linewidth]{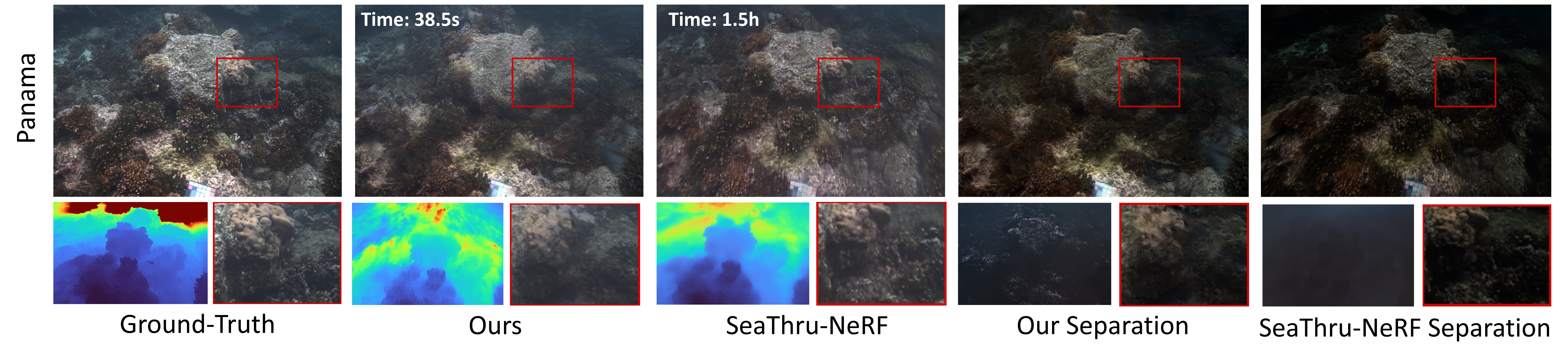}
	\caption{\textbf{Scene Separation, on ‘Panama’.} In the bottom left corner, the depth map and medium image are shown, while the bottom right corner displays a magnified view of the details. 
	}
\end{figure*}

\begin{table*}[!htbp]
	\centering
	\caption{\raggedright\textbf{Dataset Quantitative Results.} We evaluate the models based on PSNR\textsuperscript{$\uparrow$}, SSIM\textsuperscript{$\uparrow$}, LPIPS\textsuperscript{$\downarrow$}, and Avg. Time\textsuperscript{$\downarrow$}.
	}
	\renewcommand{\arraystretch}{1.2}
	\begin{tabularx}{\textwidth}{c|>{\centering\arraybackslash}X >{\centering\arraybackslash}X >{\centering\arraybackslash}X|>{\centering\arraybackslash}X >{\centering\arraybackslash}X >{\centering\arraybackslash}X|>{\centering\arraybackslash}X >{\centering\arraybackslash}X >{\centering\arraybackslash}X|>{\centering\arraybackslash}X >{\centering\arraybackslash}X >{\centering\arraybackslash}X|>{\centering\arraybackslash}X}
		\textbf{Scene}
		& \multicolumn{3}{c|}{\textbf{Cura\c{c}ao}} 
		& \multicolumn{3}{c|}{\textbf{IUI3-RedSea}} 
		& \multicolumn{3}{c|}{\textbf{JpnGradens-RedSea}} 
		& \multicolumn{3}{c|}{\textbf{Panama}} 
		& Avg.\\
		\textbf{Method}
		& PSNR & SSIM & LPIPS  
		& PSNR & SSIM & LPIPS  
		& PSNR & SSIM & LPIPS  
		& PSNR & SSIM & LPIPS 
		& Time \\ \hline
		COLMAP + SeaThru-NeRF & 24.030 & 0.667 & 0.398 & 23.476 & 0.636 & 0.351 & 23.527 & 0.631 & 0.377 & 24.167 & 0.653 & 0.381 & 1.5h \\ 
		COLMAP + 3DGS & 22.342  & 0.632 & 0.358 & 22.817 & 0.631 & 0.349& 20.411 & 0.635 & 0.310 & 22.050 & 0.651 & 0.301 & 12.2m \\ \hline
		InstantSplat & 21.031 & 0.573 & 0.419 & 20.116 & 0.571 & 0.473 & 19.421 & 0.578 & 0.419 &  22.852 &  0.642 & 0.383 & 41.4s \\ \hline
		\textbf{Ours-S} & 22.922 & 0.546 & 0.465 & 21.390 & 0.553 & 0.462 & 22.787 & 0.541 & 0.477 & 23.103 & 0.582 & 0.448 & 
		\cellcolor{lightred}23.2s\\
		\textbf{Ours-M} & \cellcolor{lightred}24.461 & \cellcolor{lightred}0.671 & \cellcolor{lightred}0.329 & \cellcolor{lightred}23.741 & \cellcolor{lightred}0.683 & \cellcolor{lightred}0.337 & \cellcolor{lightred}23.849 & \cellcolor{lightred}0.642 & \cellcolor{lightred}0.301 & \cellcolor{lightred}24.613 & \cellcolor{lightred}0.672 & \cellcolor{lightred}0.289 & 37.7s \\
	\end{tabularx}
\end{table*}

\vspace{-5pt}\noindent
\textbf{Robot Experiment Results. }In the first robotics experiment, we aimed to reconstruct a black obstacle in the tank (the first part of Fig. 1). The quantitative results are shown in TABLE II. For a simple underwater object, our model achieved excellent performance in an extremely short time, surpassing COLMAP + SeaThru-NeRF.
\begin{table}[H]
	\centering
	\caption{\raggedright\textbf{Robot Experiment Quantitative Results. }}
	\renewcommand{\arraystretch}{1.2}
	\begin{tabularx}{\columnwidth}{c|>{\centering\arraybackslash}X >{\centering\arraybackslash}X >{\centering\arraybackslash}X>{\centering\arraybackslash}X }
		\textbf{Scene}
		& \multicolumn{4}{c}{\textbf{black barricade}} \\ 
		\textbf{Method}
		& PSNR & SSIM & LPIPS & Time\\ \hline
		\textbf{Ours-M} & \cellcolor{lightred}29.512 & \cellcolor{lightred}0.953 & \cellcolor{lightred}0.243 & \cellcolor{lightred}42.3s \\   \hline
		COLMAP + SeaThru-NeRF & 28.592 & 0.912 & 0.261 &  1.5h \\
	\end{tabularx}
\end{table}

The qualitative results are presented in Fig. 5. From a visual inspection, our medium and object separation results are comparable to those of SeaThru-NeRF, both producing outstanding outcomes. 

In the second robotics experiment, the scene we aimed to reconstruct is shown in Fig. 2. It is important to note that due to variations in water quality and lighting conditions, the images obtained in this trial had lower resolution and fewer feature points, making it impossible to use COLMAP for initial camera pose estimation. As a result, 3D reconstruction methods based on SfM, including SeaThru-NeRF and 3DGS, could not proceed smoothly. However, with AquaGS, whether using 3 images or 12, we were able to quickly achieve a satisfactory result, as shown in Figure 6. This demonstrates that our SfM-Free approach has a significant advantage in underwater environments.

\vspace{-5pt}
\section{Conclusion}
\vspace{-5pt}
In this paper, we proposed a fast and stable SfM-Free underwater 3D reconstruction method called AquaGS. We tackled the limitations of SfM-based methods in underwater scenes by using a pre-trained neural network optimized with a confidence-based downsampling strategy, enabling efficient Gaussians initialization within seconds. Inspired by the SeaThru model, we addressed the medium effects overlooked by mainstream methods, utilizing Gaussians for object geometry and an MLP to capture translucent medium information, effectively separating objects from the surrounding medium. We conducted model comparisons on the standard benchmark dataset SeaThru-NeRF and our own robots. Even with low-resolution and sparse input images, AquaGS achieved excellent results in seconds. Our work significantly advanced the practical application of 3D reconstruction algorithms on robotic platforms. Future directions include further reducing the computational memory and storage size of AquaGS to facilitate its development for embedded systems.

\begin{figure*}[!htbp] 
	\centering
	\includegraphics[width=\linewidth]{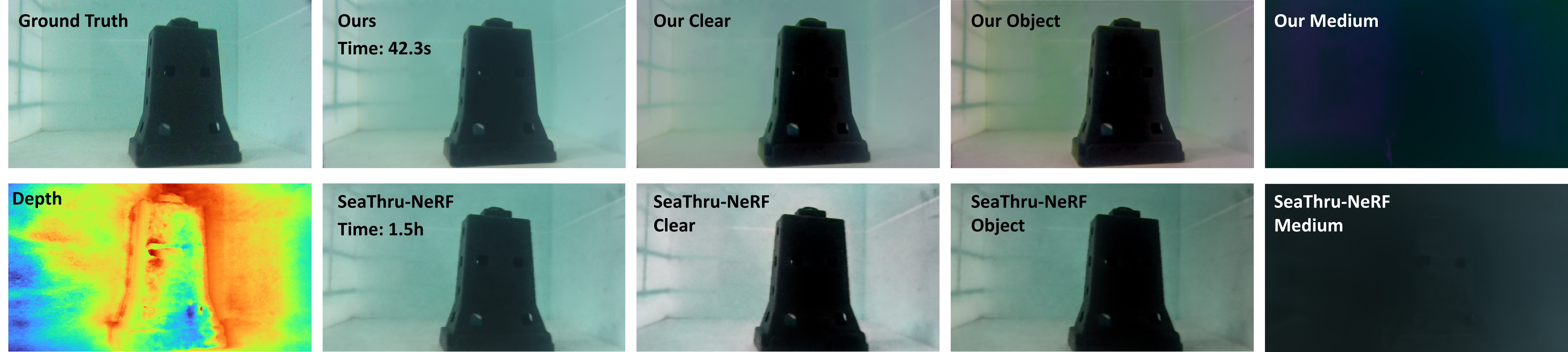}
	\caption{\textbf{Robot Experiment Results on ‘Black Barricade’.} Upon visual examination, our method for separating the medium and objects matches the performance of SeaThru-NeRF, with both approaches delivering exceptional outcomes.
	\vspace{-6pt}
	}
\end{figure*}

\begin{figure*}[htbp]  
	\centering
	\includegraphics[width=\linewidth]{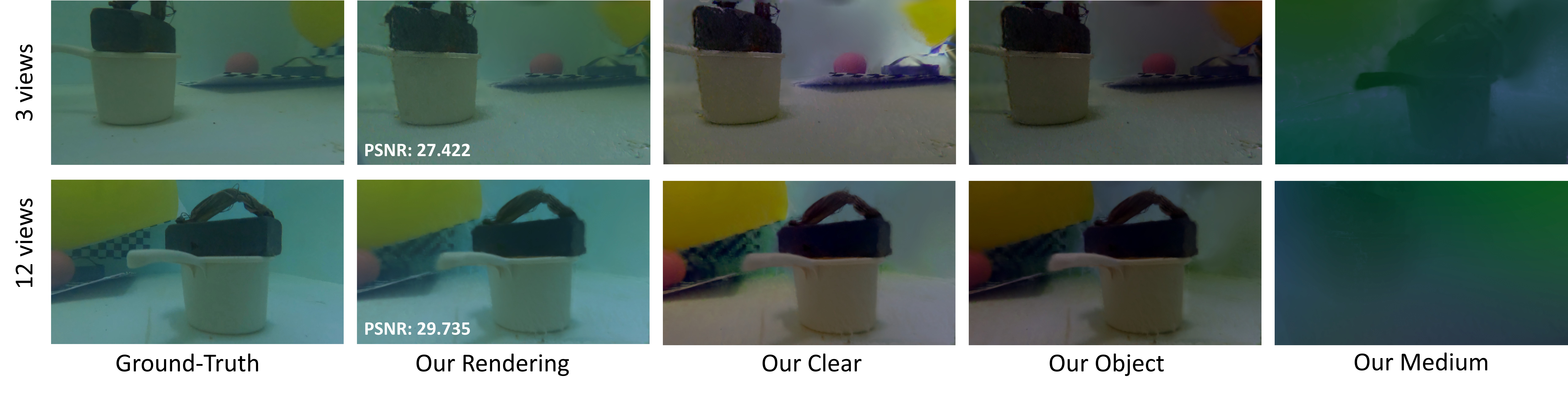}
	\caption{\textbf{Robot Experiment Results on ‘Water Scoop’. }In cases where the input images are of low quality and limited in number, SfM-based 3D reconstruction algorithms are prone to failure, whereas our algorithm demonstrates excellent robustness.
	}
	\vspace{-10pt}
\end{figure*}
\bibliographystyle{ieeetr}
\bibliography{Library.bib}
\end{document}